\documentclass[pdflatex,sn-basic,Numbered,iicol]{sn-jnl}
\usepackage[utf8]{inputenc}
\usepackage[T1]{fontenc}
\usepackage{lmodern}
\usepackage{amsmath,amssymb,amsfonts}
\usepackage{graphicx,booktabs,array,calc}
\usepackage{xcolor}
\usepackage{microtype}
\usepackage{xurl}
\usepackage{longtable,placeins}
\makeatletter
\long\def\LT@makecaption#1#2#3{%
  \LT@mcol\LT@cols c{\hbox to\z@{\hss%
    \parbox[t]\LTcapwidth{%
      \ifx#1\@gobble
        {\tablecaptionfont\raggedright#3\par}%
      \else
        \@tablecaption{#2}{#3}%
      \fi
      \vskip\belowcaptionskip}%
    \hss}}}
\makeatother
\makeatletter
\newcommand{\continuedfigurecaption}[2]{%
  \par\vskip\abovecaptionskip
  \noindent\parbox{\linewidth}{\@figurecaption{\figurename\space#1}{#2}}}
\makeatother

\makeatletter
\let\supplementtitle\title
\let\supplementauthor\author
\let\supplementmaketitle\maketitle
\let\supplement@maketitle\@maketitle
\makeatother
\setcitestyle{numbers,square,comma}
\hypersetup{hidelinks,hypertexnames=false}
\renewcommand{\arraystretch}{1.15}
\begin{document}
\title{Toward Evidence-Informed LLM Selection for Generative Agent-Based Models: Evaluating Behavioral Relationships in Epidemiology}
\author*[1]{\fnm{Sophia} \sur{Mao}}\email{smao@uic.edu}
\author[1,2]{\fnm{Keshav} \sur{Gandhi}}
\author[1]{\fnm{Alexander} \sur{Gutfraind}}
\affil*[1]{\orgdiv{Division of Epidemiology and Biostatistics,
School of Public Health},
\orgname{University of Illinois Chicago},
\orgaddress{\city{Chicago}, \state{IL}, \postcode{60612},
\country{USA}}}

\affil[2]{\orgdiv{Program for Experimental and Theoretical Modeling,
Division of Hepatology, Department of Medicine},
\orgname{Loyola University Chicago},
\orgaddress{\city{Maywood}, \state{IL}, \postcode{60153},
\country{USA}}}
\makeatletter\def\@presentaddresstxt{}\makeatother
\presentaddress{ORCID iDs: Sophia~Mao~\href{https://orcid.org/0009-0006-3519-4265}{0009-0006-3519-4265};
Keshav~Gandhi~\href{https://orcid.org/0009-0000-1718-1862}{0009-0000-1718-1862};
Alexander~Gutfraind~\href{https://orcid.org/0000-0002-3324-2220}{0000-0002-3324-2220}}

\abstract{
Representing individual-level behaviors and their relationships with social conditions is challenging in agent-based models (ABMs) when empirical data are limited. Large language models (LLMs) have been used to drive behaviors in generative ABMs, in which multiple agents can share the same underlying LLM. However, selecting an LLM for this role requires considering whether its generated behaviors reproduce established epidemiological relationships. We develop a framework for evaluating candidate LLMs, using people who inject drugs as a case study, and assess three LLMs (Claude Sonnet 4.6, DeepSeek-V3.2, and GPT-OSS-120B) against 28 literature-informed association directions drawn primarily from meta-analyses and cohort studies. We used bootstrap resampling of profiles to assess whether each relationship's agreement with its literature-informed direction was sensitive to which profiles were included. Unconditional directional recovery was 46--68\%, compared with 82--93\% agreement in knowledge probes asking LLMs to classify the same associations. Two of the three LLMs generated constant values across profiles for some behavioral outcomes. Housing-related differences had different signs between prompt versions for 7--43\% of profile--outcome pairs. Our framework identifies model- and relationship-specific limitations that could inform LLM selection and flag behaviors warranting further evaluation or alternative specification.

}

\maketitle
\clearpage

\section{Introduction}

Large language models (LLMs) have introduced a new possibility for epidemiological modeling: generating individual-level behavioral outputs that could supplement limited empirical data. This approach may be particularly useful for agent-based models (ABMs), which represent populations of simulated agents with individual characteristics and behaviors that require numerous parameters informed by empirical data and/or literature-derived estimates \cite{main-ref31}. However, empirically characterizing behaviors and contact structures can be complicated by reporting biases, and collecting such data at the population level may be costly or impractical \cite{main-ref13}. These challenges are especially pronounced for populations with limited behavioral, biological, and network data \cite{main-ref17}, making it difficult to model how human behaviors vary with individual and social factors.

One emerging application is generative agent-based modeling (GABM), which uses LLMs to generate agent behaviors and interactions within simulated environments rather than relying exclusively on predefined rules. In GABMs, a single LLM can generate behaviors across agents, with
outputs conditioned on each agent's individual characteristics and
context \cite{main-ref24}. This approach may be particularly useful when extending ABMs to behavioral decisions or contexts not captured by existing behavioral rules \cite{main-ref24,main-ref2}. For example, in an early application of GABMs to epidemiology, Williams et al. used GPT-3.5-turbo-0301 to generate context-dependent behavioral decisions, producing epidemic patterns qualitatively resembling those observed in previous epidemics \cite{main-ref34}.

Nonetheless, plausible aggregate simulation outcomes do not necessarily establish that the underlying behaviors reproduce established epidemiological relationships. Related research highlights the limitations of evaluating synthetic data based on aggregate similarity alone. For example, Nasution et al.~\cite{main-ref19} found that some relationships differed between synthetic health data and original data despite similar distributions. Rafikova and Voronin~\cite{main-ref30} also found that LLM-generated survey responses approximated aggregate human attitudes but did not fully capture variation in those attitudes. For GABMs, this raises a broader question of how epidemiological evidence can be used to evaluate LLMs and inform their selection for specific simulation applications. This is particularly relevant for people who inject drugs (PWID), among whom the transmission of bloodborne viruses such as hepatitis C virus (HCV) and human immunodeficiency virus (HIV) is shaped by interconnected behavioral, social, and contextual factors, including injection practices and housing instability \cite{main-ref4}.

Accordingly, we propose an evaluation framework to inform LLM selection for GABMs by assessing behavioral relationships against epidemiological evidence. Using PWID as a case study, we evaluate relationships generated by three LLMs against literature-informed association directions drawn primarily from meta-analyses and cohort studies. We also conduct two supplementary quantitative comparisons and assess sensitivity to prompt wording. Together, these analyses identify model- and relationship-specific limitations that could inform LLM selection for GABMs and highlight behaviors warranting further evaluation or alternative specification.

\section{Methods}

\subsection{Paired within-individual design}
We used a within-individual design in which the same individual profiles were evaluated under two housing conditions. Individual profiles were derived from the enhanced Community Outreach Intervention Projects Needle Exchange Program (CNEP+) population of PWID in Chicago \cite{main-ref17}. 224 unique agents were selected through stratified random sampling by sex, race/ethnicity, and injection experience (\textless3 vs. \ensuremath{\geq}3 years), ensuring equal representation across predefined strata. Sample inclusion criteria were age \ensuremath{\geq}18 years, HCV-susceptible status at baseline, non-zero injection frequency, and at least two recorded time points in the longitudinal CNEP+ dataset. 

Each profile was assigned two housing conditions: stable for at least one year vs. homeless for at least one month. These thresholds were chosen to distinguish long-term housing stability from current or recent homelessness. All other baseline attributes were held constant across housing conditions. The prompt scenario specified a 6-month period in 2026 during which individuals were aware that illicitly manufactured fentanyl was present in the drug supply.  LLMs were instructed to generate behavioral outcomes representing a typical 4-week period at the end of this 6-month period. The full prompts are available in the repository referenced in the Data Availability Statement.

This design yielded 448 sets of behavioral outputs per LLM (224 individuals under two housing scenarios). Prompts were executed independently. To formalize this design, let \(X_{ic}\) denote the baseline characteristics and \(Y_{ick}^{*}\) denote outcome \(k = 1,\ldots,K\) for individual profile \(i\) under housing condition \(c \in \{\text{stable}, \text{homeless}\}\). Let \(S\) denote the shared prompt context, including the scenario and task instructions. For each prompt, the LLM generated multiple behavioral outcomes (Table~\ref{tab:1}), represented as:
\begin{equation}
\label{eq:1}
\mathbf{Y}_{ic}^{*} = f_{\mathrm{LLM}}(X_{ic}, S)
= \left(Y_{ic1}^{*}, \ldots, Y_{icK}^{*}\right),
\end{equation}
where \(f_{\mathrm{LLM}}\) represents the generation process for each LLM and \(\mathbf{Y}_{ic}^{*}\) denotes the generated vector of behavioral outcomes.

\begingroup
\small
\begin{table*}[t]
\small
\centering
\caption{LLM-generated outcome variables}\label{tab:1}
\begin{tabular}{@{}
  >{\raggedright\arraybackslash}p{(\textwidth - 4\tabcolsep) * \real{0.2600}}
  >{\raggedright\arraybackslash}p{(\textwidth - 4\tabcolsep) * \real{0.5800}}
  >{\raggedright\arraybackslash}p{(\textwidth - 4\tabcolsep) * \real{0.1600}}@{}}
\toprule
\textbf{Outcome variable} &
\textbf{Definition} &
\textbf{Type} \\
\midrule
Injection frequency &
Average weekly number of injections &
Count \\

Network in-degree &
Average weekly number of unique individuals from whom the person received a used syringe &
Count \\

Network out-degree &
Average weekly number of unique individuals to whom the person gave a used syringe &
Count \\

Partners &
Average weekly number of unique injection partners &
Count \\

Non-injection routes &
Average weekly number of occasions of smoking or snorting drugs &
Count \\

Overdose &
Average weekly number of non-fatal overdoses &
Count \\

Receptive sharing &
Injections involving reuse of a syringe previously used by another person &
Proportion \\

Partners with HCV &
Injection partners who are HCV positive &
Proportion \\

Public injecting &
Injections occurring in public settings (parks, streets, shelters) &
Proportion \\

Treatment &
Enrolled in opioid use disorder treatment program &
Binary \\

HCV status &
Current HCV state, restricted to susceptible or acute infection &
Binary \\
\bottomrule
\multicolumn{3}{@{}p{\textwidth}@{}}{%
\footnotesize\textit{Note.} Variable definitions and measurements varied
across studies and did not necessarily correspond exactly to the outcome
variable definition specified by the LLM prompt. Weekly counts represent
average weekly behavior during the final four weeks of the six-month
scenario specified in the prompt.
} \\
\end{tabular}
\end{table*}
\endgroup

\subsection{Models}
Three models (Claude Sonnet 4.6, DeepSeek v3.2, and GPT-OSS-120B) were evaluated and are hereafter referred to as Claude, DeepSeek, and GPT-OSS, respectively. LLM temperature was set to zero to minimize stochastic variation for identical inputs. Thus, the analysis focuses on variation across profiles, models, and prompt wording rather than variation from repeated stochastic generations.

\subsection{Evaluation of LLM-generated relationships}

\begin{figure*}[t]
\centering
\includegraphics[width=\textwidth]{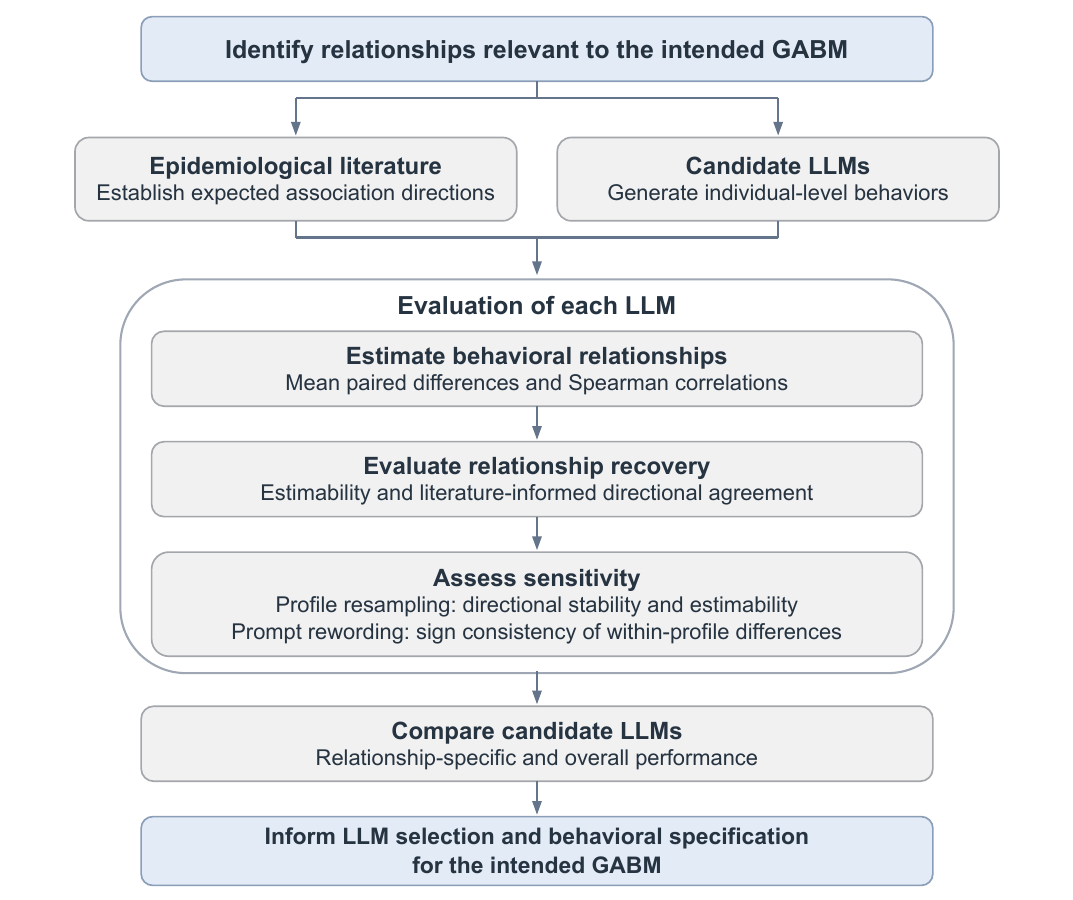}
\caption{Overview of the proposed evaluation framework for informing LLM selection in generative agent-based models. GABM, generative agent-based model; LLM, large language model}\label{fig:1}
\end{figure*}

Figure~\ref{fig:1} provides an overview of the proposed evaluation framework. The evaluation measures are considered separately rather than combined into a single score, as their relevance to LLM selection depends on the intended GABM application. The framework is intended to inform LLM selection rather than prescribe a definitive selection rule.

\subsubsection{Literature-informed directional agreement}

To identify limitations relevant to LLM selection, we first assessed agreement between LLM-generated and literature-informed association directions (Table~\ref{tab:2}; Supplementary Table~S1 in Online Resource~1). 32 literature-informed associations were identified, four of which had mixed or inconclusive prespecified directions. Although individual characteristics and contextual information can
be specified in the prompt, the implicit assumptions underlying
LLM-generated behaviors cannot be fully controlled. Association directions that are consistent across studies despite
differences in variable definitions, populations, settings, and
methodological approaches provide a basis for evaluation that is
less dependent on the specific conditions of any one study.
We therefore evaluated agreement in association directions rather
than magnitudes, retaining reported effect estimates as supporting
evidence.

\begingroup
\footnotesize
\setlength{\tabcolsep}{4pt}
\renewcommand{\arraystretch}{1.0}
\begin{table*}[!tp]
\centering
\caption{Prespecified literature-informed directions for associations between housing status and outcome variables}
\label{tab:2}
\begin{tabular}{@{}
  >{\raggedright\arraybackslash}p{(\textwidth - 4\tabcolsep) * \real{0.35}}
  >{\centering\arraybackslash}p{(\textwidth - 4\tabcolsep) * \real{0.11}}
  >{\raggedright\arraybackslash}p{(\textwidth - 4\tabcolsep) * \real{0.54}}@{}}
\toprule
\textbf{Variable pair} &
\textbf{Direction} &
\textbf{Evidence} \\
\midrule

Homelessness -- public injecting & \ensuremath{\uparrow} &
OR = 4.23 (3.17--5.65) \cite{main-khezri2025} (meta-analysis)\newline
Positive association reported \cite{main-ref4} (systematic review) \\

Homelessness -- injection frequency & \ensuremath{\uparrow} &
OR = 1.40 (1.08--1.82) \cite{main-ref6}
(multiple injections/day vs.\ less than weekly)\newline
Positive association reported \cite{main-ref4} (systematic review) \\

Homelessness -- receptive sharing & \ensuremath{\uparrow} &
Positive association reported \cite{main-ref4} (systematic review)\newline
aOR = 1.93 (1.67--2.23) \cite{main-ref20}\newline
aPR = 1.26 (1.15--1.39) \cite{main-ref21}
(cohort; any needle sharing)\newline
RR = 1.18 (0.60--2.39) \cite{main-ref27}
(cohort; partner-specific sharing) \\

Homelessness -- positive HCV status & \ensuremath{\uparrow} &
aHR = 1.86 (1.32--2.55) \cite{main-ref28}
(cohort; incident HCV)\newline
aRR = 1.64 (1.43--1.89) \cite{main-ref4}
(meta-analysis; HCV acquisition)\newline
aHR = 1.46 (1.21--1.76) \cite{main-ref14}
(cohort; incident HCV)\newline
aOR = 1.55 (1.50--1.61), currently homeless versus at risk of
homelessness among veterans \cite{main-ref7}\newline
HR = 1.20 (0.60--2.30) \cite{main-ref18}
(cohort; HCV seroconversion) \\

Homelessness -- overdose & \ensuremath{\uparrow} &
OR = 1.64 (1.49--1.81) \cite{main-ref12} (meta-analysis)\newline
aPR = 1.64 (1.49--1.79) \cite{main-ref26}\newline
aOR = 1.36 (0.70--2.63) \cite{main-ref16}
(unstable housing in prior year; overdose in prior 6 months)\newline
aOR = 1.29 (0.82--2.04) \cite{main-ref15}
(lifetime homelessness and overdose)\newline
Positive association reported \cite{main-ref4}
(systematic review)\newline
Positive association reported \cite{main-ref5}
(systematic review; fatal overdose) \\

Homelessness -- injection partners & \ensuremath{\uparrow} &
aRR = 1.48 (1.25--1.76) \cite{main-ref20}
(injection sub-network size) \\

Homelessness -- partners with HCV & \ensuremath{\leftrightarrow} &
Contextual evidence
from Arum et al.~\cite{main-ref4} and Degenhardt et al.~\cite{main-ref11}
(systematic reviews) \\

Homelessness -- treatment & \ensuremath{\leftrightarrow} &
Lower initiation and adherence to opioid substitution treatment
reported \cite{main-ref4} (systematic review)\newline
Mixed evidence for receipt and completion of substance-use treatment
\cite{main-ref5} (systematic review) \\

Homelessness -- non-injection routes & \ensuremath{\downarrow} &
aPR = 0.84 (0.83--0.85) for preference for smoking or snorting
over injecting \cite{main-ref22}
(homeless vs.\ independent living at treatment admission)\newline
Indirect evidence on homelessness and transition toward injecting
\cite{main-ref32} (systematic review) \\

\bottomrule
\multicolumn{3}{@{}p{\textwidth}@{}}{%
\footnotesize\textit{Note.}
\(\leftrightarrow\) indicates no fixed directional benchmark;
these pairs are excluded from directional scoring.
Intervals accompanying estimates are 95\% confidence intervals.
OR = odds ratio; RR = risk ratio; HR = hazard ratio;
PR = prevalence ratio. The prefix `a' denotes adjustment;
aRR in Hotton et al.~\cite{main-ref20} denotes an adjusted rate ratio.
} \\
\end{tabular}
\end{table*}
\endgroup
For profile \(i\) and outcome \(k\), the LLM-generated housing condition difference was defined as
\begin{equation}
\label{eq:2}
\Delta_{ik} =
Y_{i,\mathrm{homeless},k}^{*}
-
Y_{i,\mathrm{stable},k}^{*}.
\end{equation}

Housing-related differences defined in Equation~\ref{eq:2} were summarized using mean paired differences, expressed in percentage points for binary and proportion outcomes. A zero mean paired difference indicates no average difference between housing conditions across profiles, but does not imply that all within-profile differences are zero. Directional agreement with literature-informed associations was assessed using the signs of the mean paired differences, with zero mean paired differences classified as directional nonagreement regardless of whether the literature specified a positive or negative association. 

Associations between generated outcome variables were also evaluated using Spearman correlations, with directional agreement assessed based on the signs of the correlation coefficients. 

\subsubsection{Estimability and directional recovery}
Estimability was incorporated into the evaluation framework to account for invariant LLM-generated outcomes. An outcome was considered invariant within a housing condition when it showed no variation across individual profiles in that condition. Correlations were calculated by pooling observations across both housing conditions and were considered non-estimable if fewer than three pairwise-complete observations were available, or if either variable had no variation in the pooled sample. Invariance within one housing condition therefore did not necessarily render a pooled correlation non-estimable.

For model \(m\) and literature-informed target \(g\), let \(E_{mg}\) indicate whether the corresponding generated association is estimable (0/1) and let \(A_{mg}\) indicate whether its estimated direction matches the literature-informed expectation (0/1). For \(G\) targets with prespecified directional expectations,
\begin{equation}
\label{eq:3}
C_{m} =
\frac{\displaystyle\sum_{g=1}^{G}E_{mg}A_{mg}}
     {\displaystyle\sum_{g=1}^{G}E_{mg}},
\qquad
U_{m} = \frac{1}{G}\sum_{g=1}^{G}E_{mg}A_{mg}.
\end{equation}

In Equation~\ref{eq:3}, \(C_{m}\) measures directional recovery conditional on estimability, whereas \(U_{m}\) measures directional recovery across all prespecified targets, retaining non-estimable targets in the denominator. Reporting both measures is important because excluding non-estimable associations from conditional directional recovery could make an LLM appear more reliable than its overall performance warrants. The four targets with mixed directional findings in the literature were excluded from the directional recovery calculations.

 \subsubsection{Bootstrap directional stability}
We used stratified bootstrap resampling to assess the stability of literature-informed directional agreement. We generated 1,000 bootstrap samples by drawing profiles with replacement within the 16 original strata, preserving the original sample size in each stratum. Both housing-condition outputs were kept paired for each selected profile, and the same profile selections were used across all LLMs and prompt versions. For each of the 1,000 bootstrap samples, we recalculated housing-related mean paired differences and Spearman correlations. For each relationship, we reported the proportion of samples in which it was estimable and, among estimable samples, the proportion whose direction agreed with the literature-informed expectation.

\subsection{Supplementary quantitative comparisons}
Because the literature-informed evaluation assessed association directions rather than magnitudes, we conducted two supplementary comparisons to examine quantitative agreement beyond association directions. These comparisons were more limited in scope: the CNEP+ analysis used simulated follow-up records from an empirically informed population and covered only a subset of behavioral relationships, whereas the empirical survey analysis evaluated LLM-generated risk scores rather than behavioral outputs using a limited sample of eligible survey participants.

\subsubsection{Comparison with CNEP+ correlations}
We calculated Spearman correlations among five shared variables in the LLM-generated outputs and CNEP+ follow-up data, then compared the corresponding correlations. Agreement was summarized using mean absolute difference (MAD), with lower values indicating closer agreement. Although follow-up records were selected based on proximity to six months after baseline, the limited availability of records near this time point resulted in a median selected follow-up time of 3.49 years (interquartile range 1.59--6.19 years). Further details and sensitivity analyses are provided in Online Resource 1.

\subsubsection{Comparison with NYC HANES correlations}
We conducted a separate analysis using the 2004 New York City Health and Nutrition Examination Survey (NYC HANES) \cite{main-ref29}. The sample included 35 HCV antibody-positive and 35 HCV antibody-negative participants. LLMs were asked to generate an HCV risk score from 0 to 100 for each individual profile, with HCV antibody status excluded from the prompt. Across participants, we calculated Spearman correlations between each profile characteristic and LLM-generated HCV risk scores and, separately, between each characteristic and observed HCV antibody status. Differences between the two sets of correlations were summarized using MAD. The ability of risk scores to distinguish HCV antibody-positive from antibody-negative participants was also evaluated using the area under the receiver operating characteristic curve (AUC). Further methodological details are provided in Online Resource 1.

\subsection{Sensitivity analyses}
LLM-generated responses may vary with model choice and prompt wording. We therefore examined two sources of variation: (1) differences across LLMs, assessed by comparing their evaluation results, and (2) sensitivity to prompt wording, assessed by comparing the directions of within-profile housing-related differences between the original and paraphrased prompts, with the latter intended to preserve the same meaning.

\subsection{Knowledge probes}
As a separate diagnostic, knowledge probes prompted the LLMs to identify the expected association directions between prespecified variable pairs. These responses were compared with the corresponding literature-informed expectations (Table~\ref{tab:2} and Supplementary Table~S1). Knowledge probes assessed whether LLMs could directly identify epidemiological relationships, in contrast to the behavioral generation task, which assessed whether generated outputs reproduced those relationships. We also assessed the sensitivity of knowledge-probe responses to prompt wording by comparing responses between the original and paraphrased prompts.

\subsection{Use of generative artificial intelligence}
Claude (Anthropic), ChatGPT (OpenAI), and Codex (OpenAI) were used in an assistive capacity to develop code for analyses and figures and to suggest revisions to the manuscript. The authors made the final methodological, interpretive, and editorial decisions and take full responsibility for the work.

\section{Results}

Unless otherwise specified, results are reported for the original prompts.

\subsection{Literature-informed directional recovery}
Of the 28 literature-informed directional targets, 22 were estimable for Claude, 18 for DeepSeek, and all 28 for GPT-OSS. Conditional directional recovery was highest for Claude (81.8\%), followed by DeepSeek (72.2\%) and GPT-OSS (67.9\%). When non-estimable targets were included in the denominator, unconditional recovery was highest for GPT-OSS (67.9\%), followed by Claude (64.3\%) and DeepSeek (46.4\%). Knowledge-probe agreement exceeded both conditional and unconditional directional recovery for all three LLMs, ranging from 82.1\% to 92.9\% (Table~\ref{tab:3}).

\begin{table*}[t]
\small
\centering
\caption{Summary of evaluation results across LLMs}\label{tab:3}
\begin{tabular}{@{}
  >{\raggedright\arraybackslash}p{(\textwidth - 6\tabcolsep) * \real{0.3300}}
  >{\raggedright\arraybackslash}p{(\textwidth - 6\tabcolsep) * \real{0.2300}}
  >{\raggedright\arraybackslash}p{(\textwidth - 6\tabcolsep) * \real{0.2200}}
  >{\raggedright\arraybackslash}p{(\textwidth - 6\tabcolsep) * \real{0.2200}}@{}}
\toprule
\textbf{Metric} &
\textbf{Claude Sonnet 4.6} &
\textbf{DeepSeek v3.2} &
\textbf{GPT-OSS-120B} \\
\midrule

\multicolumn{4}{@{}l}{\textbf{Literature-informed evaluation}} \\
Estimable targets &
22/28 (78.6\%) &
18/28 (64.3\%) &
28/28 (100.0\%) \\

Conditional directional recovery &
18/22 (81.8\%) &
13/18 (72.2\%) &
19/28 (67.9\%) \\

Unconditional directional recovery &
18/28 (64.3\%) &
13/28 (46.4\%) &
19/28 (67.9\%) \\

\addlinespace
\multicolumn{4}{@{}l}{\textbf{Prompt sensitivity}} \\
Housing-difference sign agreement &
1,670/2,016 (82.8\%) &
1,881/2,016 (93.3\%) &
1,155/2,016 (57.3\%) \\

\addlinespace
\multicolumn{4}{@{}l}{\textbf{Supplementary quantitative comparisons}} \\
CNEP+ correlation MAD &
0.11 (0.08--0.14) &
0.13 (0.10--0.17) &
0.12 (0.10--0.16) \\

NYC HANES correlation MAD &
0.17 (0.11--0.27) &
0.15 (0.10--0.26) &
0.18 (0.11--0.28) \\

NYC HANES HCV risk-score AUC &
0.85 (0.76--0.93) &
0.83 (0.73--0.91) &
0.78 (0.67--0.88) \\

\addlinespace
\multicolumn{4}{@{}l}{\textbf{Knowledge probes}} \\
Literature-informed directional agreement &
26/28 (92.9\%) &
25/28 (89.3\%) &
23/28 (82.1\%) \\

Agreement between prompt versions &
31/32 (96.9\%) &
31/32 (96.9\%) &
28/32 (87.5\%) \\

\bottomrule

\multicolumn{4}{@{}p{\dimexpr\textwidth\relax}@{}}{%
\footnotesize\textit{Note.} Parentheses indicate percentages for count-based metrics or 95\% stratified bootstrap confidence intervals for MADs and AUCs.
} \\

\end{tabular}
\end{table*}

\subsection{Relationship-specific differences}
\begin{figure*}[p]
\centering
\includegraphics[width=\textwidth]{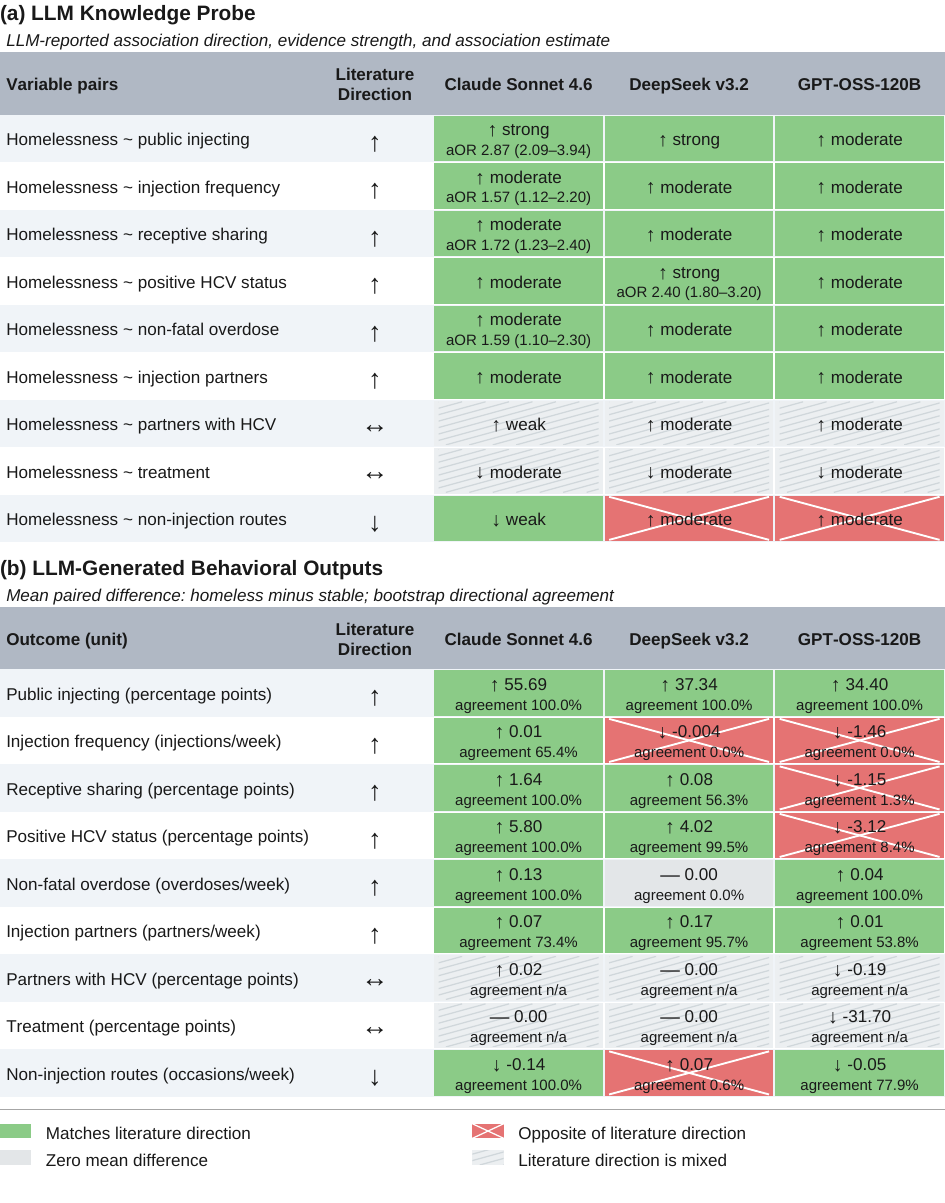}
\caption{Housing-related associations: literature-informed directions and LLM results. (a) Knowledge-probe responses; (b) LLM-generated behavioral outputs. Bootstrap agreement denotes the percentage of estimable resamples matching the literature-informed direction.}
\label{fig:2}
\end{figure*}

Public injecting was higher under homelessness than under stable housing, with differences of 55.69 percentage points for Claude, 37.34 for DeepSeek, and 34.40 for GPT-OSS, consistent with the literature-informed direction. Other housing-related differences varied by model. Receptive syringe sharing was higher by 1.64 and 0.08 percentage points for Claude and DeepSeek, respectively, but lower by 1.15 percentage points for GPT-OSS, despite all three models identifying the expected positive association in knowledge probes (Figure~\ref{fig:2}).

Differences were also observed in correlations among generated outcomes. For injection frequency and non-injection routes, the literature-informed direction was negative, which all three models correctly identified in knowledge probes. However, the generated Spearman correlations were 0.85 for Claude, -0.07 for DeepSeek, and 0.45 for GPT-OSS. Similarly, all three models generated positive correlations between non-injection routes and HCV-positive status (0.05, 0.11, and 0.10, respectively), despite correctly identifying the expected negative direction in knowledge probes (Supplementary Figure~S1).

\subsection{Invariant outcomes and zero mean paired differences}
Under both housing conditions, Claude and DeepSeek assigned no treatment to every profile. DeepSeek also assigned zero overdoses to every profile. In contrast, all GPT-OSS outcomes varied across profiles within both housing conditions.

Across the 11 outcomes evaluated by three LLMs, five of the 33 model--outcome combinations had zero mean paired differences. Three resulted from the invariant outcomes described above, while the other two involved outcomes that varied across profiles but remained unchanged within each profile across housing conditions. Under the paraphrased prompt, GPT-OSS exhibited a different pattern: the mean paired difference in HCV classifications was zero because positive and negative within-profile differences canceled out.

\subsection{Bootstrap directional agreement and estimability}
Across 1,000 stratified bootstrap resamples, the housing difference in public injecting matched the literature-informed direction in 100\% of resamples for all three LLMs. For receptive syringe sharing, directional agreement was 100\% for Claude, 56.3\% for DeepSeek, and 1.3\% for GPT-OSS.

For correlations among generated outcomes, the literature-informed negative association between injection frequency and non-injection routes was recovered in 97.9\% of bootstrap resamples for DeepSeek but in none for Claude or GPT-OSS. In contrast, the expected negative association between non-injection routes and HCV-positive status was rarely recovered by any model (16.0\% for Claude, 4.7\% for DeepSeek, and 3.7\% for GPT-OSS; Supplementary Figure~S1).

Under the original behavioral prompts, all relationships that were estimable in the full sample remained estimable across all 1,000 bootstrap resamples, while non-estimable relationships remained non-estimable.

\subsection{Sensitivity to prompt wording}
The sign of the within-profile housing difference was unchanged between the original and paraphrased behavioral prompts for 82.8\% of profile--outcome pairs for Claude, 93.3\% for DeepSeek, and 57.3\% for GPT-OSS. These comparisons covered 2,016 profile--outcome pairs per LLM (224 profiles across nine numeric and proportion outcomes). Separately, knowledge-probe direction categories were unchanged between prompt versions for 96.9\% of relationships for both Claude and DeepSeek and 87.5\% for GPT-OSS (Table~\ref{tab:3}).

\subsection{Supplementary quantitative comparisons}

MADs between LLM-generated correlations and CNEP+ simulated follow-up correlations ranged from 0.11 to 0.13 across the three models (Table~\ref{tab:3}; Supplementary Figure~S2). For the NYC HANES comparison, MADs ranged from 0.15 to 0.18. LLM-generated HCV risk scores distinguished between HCV antibody-positive and antibody-negative survey participants, with AUCs ranging from 0.78 to 0.85 (Table~\ref{tab:3}).

\section{Discussion}

Across the three evaluated LLMs, generated behaviors did not consistently reproduce literature-informed association directions, and models differed in both directional recovery and sensitivity to profile resampling and prompt wording. Differences in performance across variables have similarly been reported in previous evaluations of LLM-generated survey data \cite{main-ref3}. Although Claude achieved the highest conditional directional recovery, the model did not generate any variation in treatment status across profiles. If these outputs were used directly in a GABM, individual-level differences in treatment status would not be captured. Although GPT-OSS did not exhibit this limitation and achieved the highest unconditional directional recovery, its outputs still reflected relationships opposite to literature-informed expectations, such as lower receptive syringe sharing under homelessness than under stable housing. This discrepancy could be consequential in a GABM intended to model the relationship between housing instability and HCV transmission.

Beyond directional agreement, stability is another important consideration. Even when the expected direction was recovered in the full sample, as with receptive syringe sharing for DeepSeek, bootstrap resampling showed that recovery could depend on the composition of sampled profiles. Furthermore, the relatively high unconditional recovery of GPT-OSS did not translate into greater consistency across prompt versions; in fact, GPT-OSS showed the greatest sensitivity to prompt wording in housing-related differences. 

Notably, knowledge-probe agreement exceeded directional recovery in generated behaviors for all three LLMs. Our comparisons suggest that (1) an LLM's ability to correctly identify known epidemiological relationships does not necessarily translate into reproducing them in individual-level behavioral outputs and (2) LLM knowledge-probe responses were generally less sensitive to prompt wording than generated behavioral outputs. Knowledge-based evaluations and assessments of aggregate ABM outcomes alone may therefore overlook limitations in the underlying behavioral relationships. Conventional LLM benchmarks have largely emphasized factual knowledge and reasoning tasks \cite{main-ref8}, making this distinction particularly important.

Evaluating LLM-generated behaviors is particularly relevant for infectious disease transmission modeling, where individual behaviors and interactions can influence downstream transmission and epidemiological outcomes \cite{main-ref1,main-ref17}. Although LLMs cannot fully overcome persistent challenges related to missing data and underrepresented populations, they may be useful for modeling hard-to-study populations, such as people experiencing homelessness, and rapidly changing contexts, such as an evolving drug landscape. In GABMs where a single LLM generates behaviors across agents, our framework could inform model selection by identifying relationship-specific limitations across candidate LLMs. Rather than selecting the LLM with the highest overall directional recovery, researchers could prioritize the behavioral relationships most relevant to their intended simulation and compare candidate LLMs based on the recovery and stability of those relationships. Limitations in less relevant relationships may be more tolerable, whereas limitations in relationships central to the research question may warrant selecting a different LLM or specifying those behaviors through predefined rules or literature-informed parameters, where available. Such evaluations could also help researchers assess whether lower-cost or open-source LLMs provide adequate agreement with epidemiological evidence for the intended application.

Several limitations should be considered when interpreting these evaluations. Differences in study populations, variable definitions, and analytical approaches may limit the applicability of some literature-informed association directions. Supplementary quantitative comparisons were limited to selected variables and benchmarks and do not establish overall rankings of LLM performance. Finally, our evaluation did not assess whether the identified behavioral limitations affect downstream GABM outcomes.

Future work could assess whether LLM selection informed by the framework improves agreement between downstream GABM outcomes and empirical data, compared with LLM selection without the framework. Another direction is to investigate whether knowledge-probe performance predicts behavioral relationship recovery across a broader range of models and prompts. If such a relationship exists, knowledge probes could provide a more computationally efficient approach to screening candidate LLMs before evaluating their generated behaviors. Additionally, more extensive prompt-perturbation experiments could determine whether sensitivity to prompt wording varies across epidemiological conditions, behavioral relationships, or domains.

\section{Conclusion}

Differences in relationship recovery and consistency across the evaluated LLMs highlight the importance of relationship-specific evaluation. Our framework could help researchers evaluate LLMs based on the behavioral relationships most relevant to their intended GABMs. Literature-informed directional agreement provides a basis for this evaluation, while invariant outcomes, estimability, and sensitivity to profile resampling and prompt wording provide additional information about the limitations and consistency of relationship recovery. Whether incorporating such evidence into LLM selection improves downstream epidemiological outcomes remains to be established.

\section*{Supplementary Information}

\noindent\textbf{Online Resource 1.} Supplementary methods, Tables S1--S4, Figures S1--S3, and supplementary references.

\section*{Statements and Declarations}

\subsection*{Data Availability Statement}
The data, LLM-generated outputs, and code used in this study are
publicly available at \url{https://github.com/mayonais/hcv_llm} and \url{https://doi.org/10.5281/zenodo.24734814}.
The 2004 New York City Health and Nutrition Examination Survey data are
available from the Inter-university Consortium for Political and Social
Research \cite{main-ref29}.

\subsection*{Funding}
The work was supported in part by National Institutes of Health {[}grant
number R01GM121600{]}.

\subsection*{Competing Interests}
In addition to his affiliation with the
University of Illinois, A.G. is employed by Amazon.com but his employer
did not have any part in this study. The other authors declare no
competing interests.

\subsection*{Ethics Approval and Consent to Participate}
Not applicable. This study analyzed existing de-identified survey data
\cite{main-ref29} and simulated follow-up outcomes based on CNEP+ population \cite{main-ref17}. No new data were collected from human
participants.

\subsection*{Author Contributions}
\textbf{Sophia Mao}: Conceptualization, data curation, formal analysis,
investigation, methodology, project administration, software,
validation, visualization, writing -- original draft, writing -- review
\& editing. \textbf{Keshav Gandhi}: Writing -- review \& editing. \textbf{Alexander Gutfraind}: Conceptualization, funding acquisition,
resources, supervision, writing -- review \& editing.

\begingroup
\sloppy

\endgroup
\clearpage
\onecolumn
\newgeometry{reversemp,top=26mm,textwidth=174mm,textheight=234mm,hcentering,bindingoffset=6mm,headheight=5.5pt,headsep=5.6mm,footskip=10mm,marginparsep=5mm,marginparwidth=12mm}
\setlength{\columnsep}{5mm}
\setlength{\footnotesep}{5.75pt}
\setlength{\skip\footins}{8pt plus 4pt}
\setlength{\LTpre}{8pt}
\setlength{\LTpost}{8pt}
\setlength{\LTcapwidth}{\textwidth}
\setcounter{page}{1}
\setcounter{section}{0}
\setcounter{subsection}{0}
\setcounter{subsubsection}{0}
\setcounter{table}{0}
\setcounter{figure}{0}
\setcounter{equation}{0}
\renewcommand{\thetable}{S\arabic{table}}
\renewcommand{\thefigure}{S\arabic{figure}}
\makeatletter
\@iicolfalse
\let\title\supplementtitle
\let\author\supplementauthor
\let\maketitle\supplementmaketitle
\let\@maketitle\supplement@maketitle
\let\@abstract\@empty
\let\@keywords\@empty
\let\artauthors\@empty
\let\auaddress\@empty
\let\corrauthemail\@empty
\let\authemail\@empty
\let\authorsep\@empty
\let\sep\@empty
\aucount=0\relax
\corraucount=0\relax
\emailcnt=0\relax
\addcount=0\relax
\setcounter{affn}{0}
\makeatother
\pdfbookmark[0]{Online Resource 1: Supplementary Information}{supplement-start}
\title{Online Resource 1: Supplementary Information}
\subtitle{Toward Evidence-Informed LLM Selection for Generative Agent-Based Models: Evaluating Behavioral Relationships in Epidemiology\\[0.4em]
\normalsize Journal of Computational Social Science}
\author*[1]{\fnm{Sophia} \sur{Mao}}\email{smao@uic.edu}
\author[1,2]{\fnm{Keshav} \sur{Gandhi}}
\author[1]{\fnm{Alexander} \sur{Gutfraind}}
\affil*[1]{\orgdiv{Division of Epidemiology and Biostatistics,
School of Public Health}, \orgname{University of Illinois at Chicago},
\orgaddress{\street{1603 W. Taylor St.}, \city{Chicago},
\state{IL}, \postcode{60612}, \country{USA}}}
\affil[2]{\orgdiv{Program for Experimental and Theoretical Modeling,
Division of Hepatology, Department of Medicine},
\orgname{Loyola University Chicago},
\orgaddress{\city{Maywood}, \state{IL}, \postcode{60153}, \country{USA}}}
\makeatletter\def\@presentaddresstxt{}\makeatother
\presentaddress{ORCID iDs: Sophia~Mao~\href{https://orcid.org/0009-0006-3519-4265}{0009-0006-3519-4265};
Keshav~Gandhi~\href{https://orcid.org/0009-0000-1718-1862}{0009-0000-1718-1862};
Alexander~Gutfraind~\href{https://orcid.org/0000-0002-3324-2220}{0000-0002-3324-2220}}
\maketitle

\noindent\textbf{Abbreviations.} HCV, hepatitis C virus;
LLM, large language model; ABM, agent-based model;
CNEP+, enhanced Community Outreach Intervention Projects Needle Exchange
Program; NYC HANES, New York City Health and Nutrition Examination
Survey; MAD, mean absolute difference; AUC, area under the receiver
operating characteristic curve; CI, confidence interval; NaN, not a number.

\section{Literature-informed evaluation of generated behavioral relationships}

Table~\ref{tab:S1} presents the literature-informed directions for the 23 relationships between outcome variables, complementing the nine housing-related relationships presented in Table~2 of the main text.

For analysis, generated HCV status was coded as 0 for susceptible and 1 for acute infection, and treatment status was coded as 0 for no treatment and 1 for treatment. Public injecting, receptive syringe sharing, and the proportion of partners with HCV were expressed as proportions ranging from 0 to 1, inclusive.

\begingroup
\small
\begin{longtable}[]{@{}
  >{\raggedright\arraybackslash}p{(\columnwidth - 4\tabcolsep) * \real{0.360}}
  >{\centering\arraybackslash}p{(\columnwidth - 4\tabcolsep) * \real{0.1100}}
  >{\raggedright\arraybackslash}p{(\columnwidth - 4\tabcolsep) * \real{0.530}}@{}}
\caption{Prespecified literature-informed directions for associations between variable pairs}
\label{tab:S1}\\

\toprule
\textbf{Variable pair} &
\textbf{Direction} &
\textbf{Evidence} \\
\midrule
\endfirsthead

\toprule
\textbf{Variable pair} &
\textbf{Direction} &
\textbf{Evidence} \\
\midrule
\endhead

\bottomrule
\multicolumn{3}{@{}p{\columnwidth}@{}}{%
\footnotesize\textit{Note.}
\(\leftrightarrow\) indicates no fixed directional benchmark;
these pairs are excluded from directional scoring.
Intervals accompanying estimates are 95\% confidence intervals.
OR = odds ratio; RR = risk ratio; HR = hazard ratio;
PR = prevalence ratio; PRR = pooled risk ratio; IRR = incidence rate ratio;
OAT = opioid agonist treatment; NSP = needle and syringe programme;
aOR, aRR, aPR, aHR = adjusted measures.
For Walker et al.~\cite{supplement-walker2025},
aRR denotes an adjusted rate ratio.
Estimates reflect the source studies' own variable definitions and comparisons;
some evidence is indirect for the generated variables.
} \\
\endlastfoot

Public injecting -- positive HCV status & \ensuremath{\uparrow} &
\begin{minipage}[t]{\linewidth}\raggedright
OR = 1.55 (1.18--2.02) \cite{supplement-ref15}
(meta-analysis)\\
aOR = 1.91 (1.33--2.75) \cite{supplement-ref3}
(HCV seropositivity)\strut
\end{minipage} \\

Public injecting -- overdose & \ensuremath{\uparrow} &
OR = 2.51 (2.01--3.13) \cite{supplement-ref15}
(meta-analysis) \\

Public injecting -- receptive sharing & \ensuremath{\uparrow} &
\begin{minipage}[t]{\linewidth}\raggedright
OR = 2.41 (1.97--2.94) \cite{supplement-ref15}
(meta-analysis; needle/syringe sharing)\strut
\end{minipage} \\

Public injecting -- injection frequency & \ensuremath{\uparrow} &
OR range across studies = 1.96--5.34
\cite{supplement-ref15} (systematic review) \\

Public injecting -- treatment & \ensuremath{\downarrow} &
OR = 0.61 (0.48--0.76) for methadone maintenance treatment
(longitudinal study reported in Khezri et al.~\cite{supplement-ref15}) \\

Receptive sharing -- positive HCV status & \ensuremath{\uparrow} &
\begin{minipage}[t]{\linewidth}\raggedright
HR = 2.1 (1.2--3.7) \cite{supplement-ref9}
(cohort; HCV seroconversion)\\
PRR = 1.94 (1.53--2.46) \cite{supplement-ref20}
(meta-analysis; HCV seroconversion)\\
aHR = 1.79 (1.44--2.22) \cite{supplement-ref4}
(cohort; incident HCV)\\
Indirect evidence: aRR = 1.53 (1.01--2.31) for syringe sharing,
without distinguishing receptive from distributive sharing
(Mackesy-Amiti et al.~\cite{supplement-ref18};
baseline analysis; partner-specific sharing)\\
aPR = 1.20 (1.09--1.33) \cite{supplement-ref12}
(cohort; any needle sharing)\strut
\end{minipage} \\

Receptive sharing -- overdose & \ensuremath{\uparrow} &
\begin{minipage}[t]{\linewidth}\raggedright
Indirect evidence: aOR = 2.65 (1.43--4.94) for sharing syringes
and/or other injection paraphernalia \cite{supplement-ref11}\\
aOR = 1.70 (1.03--2.81) \cite{supplement-ref6}
(lifetime overdose)\strut
\end{minipage} \\

Treatment -- overdose & \ensuremath{\downarrow} &
OR = 0.72 (0.64--0.81) \cite{supplement-ref2}
(meta-analysis; current OAT) \\

Treatment -- positive HCV status & \ensuremath{\downarrow} &
aHR = 0.74 (0.56--0.97) \cite{supplement-ref4}
(cohort; OAT; incident HCV) \\

Treatment -- receptive sharing & \ensuremath{\downarrow} &
\begin{minipage}[t]{\linewidth}\raggedright
aOR = 0.046 (0.034--0.061) \cite{supplement-walker2025}
(OAT only vs.\ neither OAT nor NSP; sharing at last injection)\\
Negative association reported \cite{supplement-ref14}
(cohort baseline; opioid substitution treatment)\strut
\end{minipage} \\

Treatment -- non-injection routes & \ensuremath{\downarrow} &
Indirect evidence: annual decline in any non-injection opioid use after
OAT initiation; aOR = 0.89 (0.85--0.92) per year
\cite{supplement-dong2020} (cohort) \\

Treatment -- injection frequency & \ensuremath{\downarrow} &
aRR = 0.21 (0.12--0.36) \cite{supplement-walker2025}
(OAT only vs.\ neither OAT nor NSP) \\

Injection frequency -- positive HCV status & \ensuremath{\uparrow} &
\begin{minipage}[t]{\linewidth}\raggedright
HR = 2.7 (1.0--7.6) \cite{supplement-ref9}
(cohort; HCV seroconversion)\\
aOR = 1.26 (1.09--1.45) \cite{supplement-ref3}
(HCV seropositivity)\strut
\end{minipage} \\

Injection frequency -- overdose & \ensuremath{\uparrow} &
\begin{minipage}[t]{\linewidth}\raggedright
aOR = 1.55 (1.00--2.40) \cite{supplement-ref7}
(\(\geq 3\) vs.\ 1--2 injections/day; overdose in prior 6 months)\\
OR = 1.19 (0.73--1.93) \cite{supplement-ref10}
(20--30 vs.\ 1--19 injecting days/month; lifetime overdose)\\
aOR = 1.05 (1.02--1.08) \cite{supplement-ref21}
(cohort; per additional daily injection)\strut
\end{minipage} \\

Injection frequency -- receptive sharing & \ensuremath{\uparrow} &
\begin{minipage}[t]{\linewidth}\raggedright
aPR = 1.39 (1.22--1.58) \cite{supplement-ref12}
(cohort; any needle sharing)\\
OR = 1.02 (0.62--1.69) \cite{supplement-ref8}
(partner-specific sharing)\\
Positive association reported \cite{supplement-ref27}
(indirect; study-level meta-regression)\strut
\end{minipage} \\

Injection frequency -- non-injection routes & \ensuremath{\downarrow} &
\begin{minipage}[t]{\linewidth}\raggedright
Indirect evidence on transition to injecting
\cite{supplement-ref22} (systematic review)\\
Negative association reported \cite{supplement-ref16}
(cohort; smoking fentanyl)\strut
\end{minipage} \\

Non-injection routes -- positive HCV status & \ensuremath{\downarrow} &
\begin{minipage}[t]{\linewidth}\raggedright
aOR = 0.72 (0.52--0.99) \cite{supplement-desjarlais2018}
(current non-injecting heroin use; HCV seropositivity)\strut
\end{minipage} \\

Non-injection routes -- overdose & \ensuremath{\leftrightarrow} &
\begin{minipage}[t]{\linewidth}\raggedright
OR = 0.63 (0.14--1.95) \cite{supplement-ref10}
(smoking instead of injecting as a coping strategy;
overdose in prior 6 months)\strut
\end{minipage} \\

Partners -- receptive sharing & \ensuremath{\downarrow} &
\begin{minipage}[t]{\linewidth}\raggedright
OR = 0.60 (0.28--1.30) \cite{supplement-ref8}
(partner-specific sharing; indirect)\\
RR = 0.59 (0.25--1.43) \cite{supplement-ref19}
(cohort; sharing with enrolled HCV-positive partner; indirect)\\
PR = 0.49 (0.34--0.70) \cite{supplement-ref23}
(partner-specific sharing; indirect)\strut
\end{minipage} \\

Partners -- injection frequency & \ensuremath{\uparrow} &
aHR = 1.41 (1.21--1.64) \cite{supplement-ref25}
(cohort; progression in injection frequency) \\

Partners -- positive HCV status & \ensuremath{\uparrow} &
aHR = 1.05 (1.02--1.13) \cite{supplement-ref25}
(cohort; incident HCV) \\

Partners with HCV -- receptive sharing & \ensuremath{\leftrightarrow} &
\begin{minipage}[t]{\linewidth}\raggedright
PR = 1.02 (0.52--1.97) \cite{supplement-ref23}
(perceived partner HCV status; partner-specific sharing)\\
OR = 0.49 (0.25--0.95) \cite{supplement-ref8}
(perceived partner HCV status; partner-specific sharing)\strut
\end{minipage} \\

Positive HCV status -- overdose & \ensuremath{\uparrow} &
\begin{minipage}[t]{\linewidth}\raggedright
aOR = 3.24 (2.60--4.03) \cite{supplement-ref7}
(HCV antibody positivity; overdose in prior 6 months)\\
OR = 2.69 (1.98--3.66) \cite{supplement-ref3}
(HCV seropositivity; lifetime overdose)\\
aOR = 2.64 (1.47--4.74) \cite{supplement-ref6}
(HCV antibody positivity; lifetime overdose)\\
aOR = 2.44 (1.20--4.97) for chronic HCV infection
\cite{supplement-ref24}
(lifetime non-fatal opioid overdose)\strut
\end{minipage} \\

\end{longtable}
\endgroup

\FloatBarrier
\begin{figure}[p]
\centering
\includegraphics[page=1,width=\linewidth]{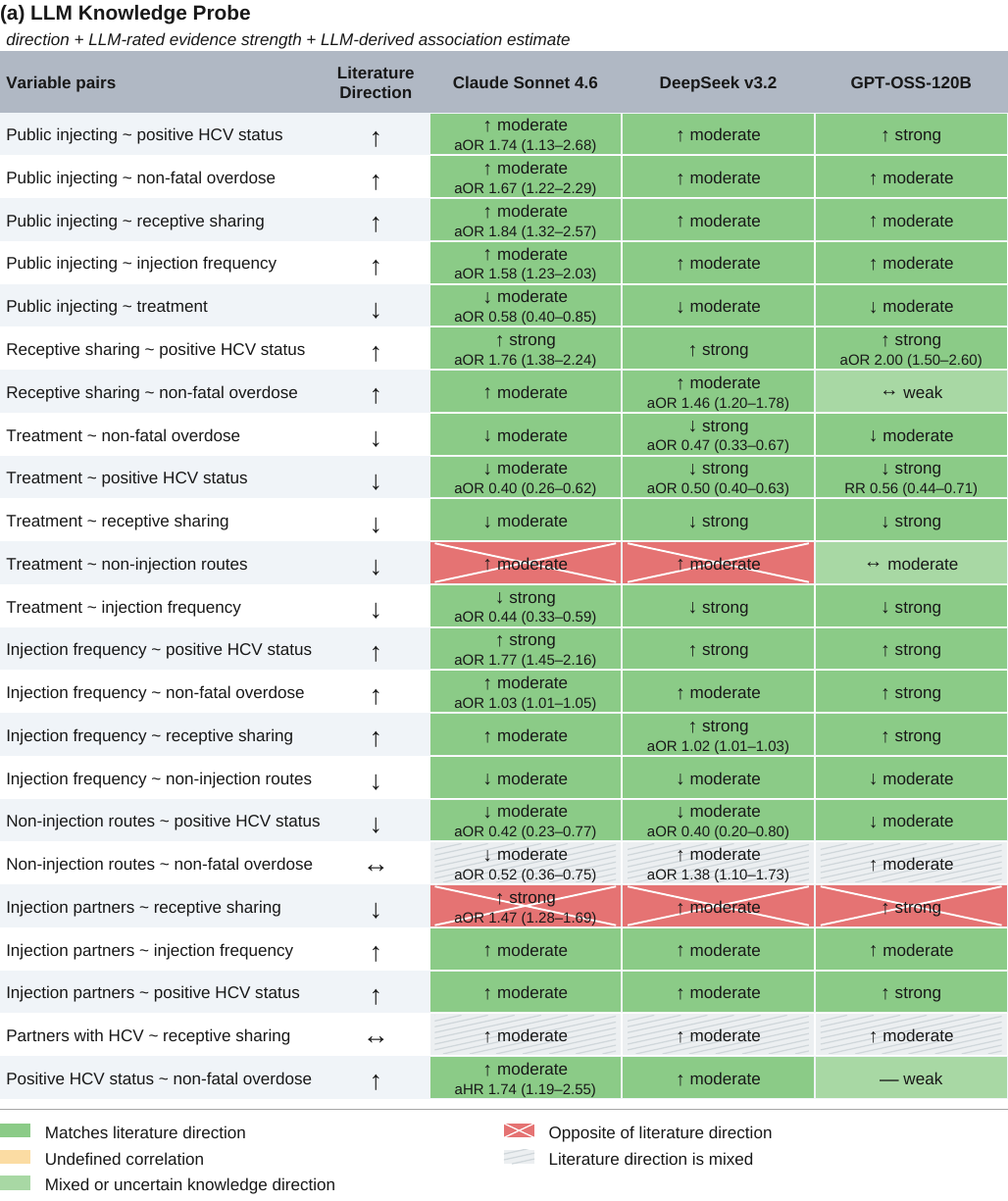}
\caption{Literature-informed directions and LLM results for associations between outcome variables.
(a) Knowledge-probe directions, LLM-rated evidence strength, and
LLM-reported association estimates and intervals, where supplied.}
\label{fig:S1}
\end{figure}

\begin{figure}[p]
\centering
\includegraphics[page=2,width=\linewidth]{FigS1.pdf}
\continuedfigurecaption{S1}{Literature-informed directions and LLM results for associations between outcome variables (continued).
(b) Spearman correlations pooled across housing conditions.
Bootstrap agreement denotes the percentage of estimable resamples
matching the literature-informed direction.
}
\end{figure}
\FloatBarrier

\section{Invariant outcomes and zero mean paired differences}

At least two nonmissing observations were required to assess invariance within a housing condition. Paired differences were calculated only for profiles with observed outcomes under both housing conditions.

Under the paraphrased behavioral prompt, one profile was excluded from the GPT-OSS HCV status comparison due to a missing value, leaving 223 profiles. Nineteen profiles changed from susceptible to acute between stable housing and homelessness, and 19 changed in the opposite direction, resulting in a zero mean paired difference.

\section{Bootstrap directional agreement and estimability}

Under both the original and paraphrased prompts, 17 of 23 correlations
were estimable in the full sample for Claude, 12 for DeepSeek, and all
23 for GPT-OSS. These correlations remained estimable in every bootstrap
resample, except for six Claude correlations involving HCV status
under the paraphrased prompt, which were estimable in 876 of 1,000
resamples (87.6\%). Non-estimable correlations remained non-estimable
in all resamples under both prompt versions. All nine mean paired
housing differences were estimable in every resample across models
and prompt versions.

\section{Sensitivity to behavioral prompt wording}

Let \(\Delta_{imk}^{(\mathrm{orig})}\) and
\(\Delta_{imk}^{(\mathrm{para})}\) denote the within-profile housing
differences (Equation~2 of the main text) for profile \(i\), model \(m\),
and outcome \(k\) under the original and paraphrased prompts,
respectively. The proportion of matching signs was
\[
Q_{mk}
=\frac{1}{N}\sum_{i=1}^{N}
\mathbf{1}\!\left\{
\operatorname{sign}\!\left(\Delta_{imk}^{(\mathrm{orig})}\right)
=\operatorname{sign}\!\left(\Delta_{imk}^{(\mathrm{para})}\right)
\right\},
\qquad \operatorname{sign}(0)=0.
\]
All 224 profiles had complete observations under both housing
conditions and prompt versions for the nine numeric and proportion
outcomes (\(N=224\)). The model-level summary was the mean of \(Q_{mk}\) across
these outcomes, equivalent to the proportion of matching signs
across 2,016 profile--outcome pairs per LLM. Outcome-specific
agreement ranged from 32.1\% to 100\% (Table~\ref{tab:S2}).

\begingroup
\small
\begin{longtable}[]{@{}
  >{\raggedright\arraybackslash}p{(\columnwidth - 6\tabcolsep) * \real{0.3518}}
  >{\raggedright\arraybackslash}p{(\columnwidth - 6\tabcolsep) * \real{0.2277}}
  >{\raggedright\arraybackslash}p{(\columnwidth - 6\tabcolsep) * \real{0.2334}}
  >{\raggedright\arraybackslash}p{(\columnwidth - 6\tabcolsep) * \real{0.1871}}@{}}
\caption{Outcome-specific agreement (\%) in within-profile housing-difference signs between the original and paraphrased prompts across 224 profiles per model}\label{tab:S2}\\
\toprule\noalign{}
\begin{minipage}[b]{\linewidth}\raggedright
\textbf{Outcome}
\end{minipage} & \begin{minipage}[b]{\linewidth}\raggedright
\textbf{Claude Sonnet 4.6}
\end{minipage} & \begin{minipage}[b]{\linewidth}\raggedright
\textbf{DeepSeek v3.2}
\end{minipage} & \begin{minipage}[b]{\linewidth}\raggedright
\textbf{GPT-OSS-120B}
\end{minipage} \\
\midrule\noalign{}
\endfirsthead
\toprule\noalign{}
\begin{minipage}[b]{\linewidth}\raggedright
\textbf{Outcome}
\end{minipage} & \begin{minipage}[b]{\linewidth}\raggedright
\textbf{Claude Sonnet 4.6}
\end{minipage} & \begin{minipage}[b]{\linewidth}\raggedright
\textbf{DeepSeek v3.2}
\end{minipage} & \begin{minipage}[b]{\linewidth}\raggedright
\textbf{GPT-OSS-120B}
\end{minipage} \\
\midrule\noalign{}
\endhead
\bottomrule\noalign{}
\endlastfoot
Overdose & 49.1 & 100.0 & 32.1 \\
Public injecting & 100.0 & 74.6 & 98.7 \\
Receptive sharing & 69.6 & 94.6 & 40.2 \\
Injection partners & 71.4 & 90.6 & 35.3 \\
Network in-degree & 97.8 & 96.9 & 71.9 \\
Network out-degree & 97.3 & 99.1 & 67.4 \\
Partner HCV proportion & 97.8 & 100.0 & 88.4 \\
Injection frequency & 97.3 & 99.6 & 33.5 \\
Non-injection routes & 65.2 & 84.4 & 48.2 \\
\end{longtable}
\endgroup

\FloatBarrier
\section{Supplementary quantitative comparisons}

\subsection{CNEP+ simulated follow-up data}

\begin{figure}[htbp]
\centering
\includegraphics[width=\linewidth,height=205mm,keepaspectratio]{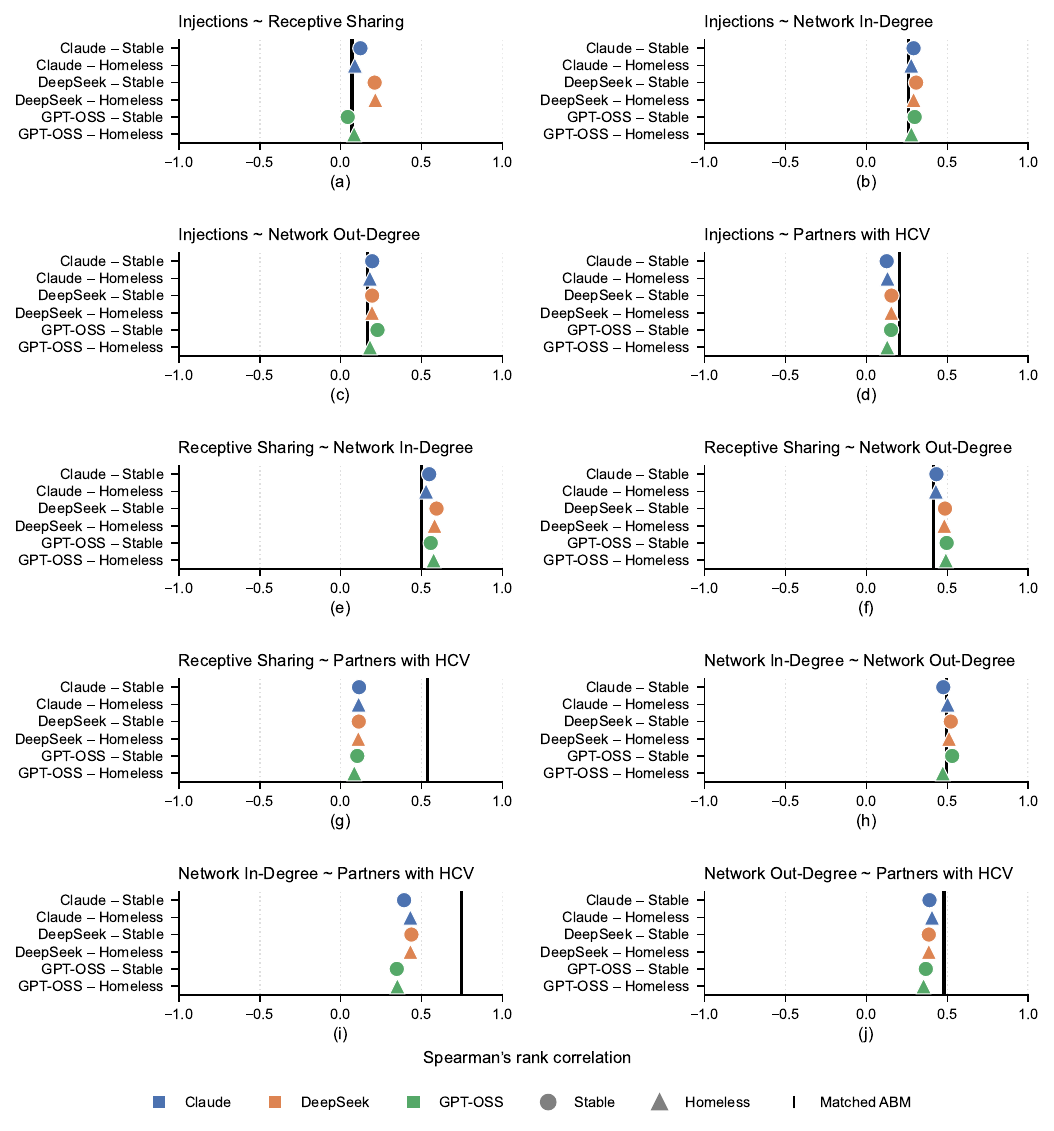}
\caption{Comparison of LLM-generated correlations with CNEP+ simulated follow-up correlations. Panels (a)--(j) show
the Spearman correlation for each pair of the five shared variables
in LLM-generated outputs under stable housing (circles) and
homelessness (triangles) for each model (colors). The vertical line
denotes the corresponding CNEP+ correlation.}
\label{fig:S2}
\end{figure}

The five shared variables were injection frequency, receptive syringe
sharing, network in-degree, network out-degree, and the proportion
of partners with HCV. Each agent's CNEP+ follow-up record was matched to its LLM-generated
outputs under both housing conditions. MAD was calculated using
pooled correlations (Table~3 of the main text), whereas
Figure~\ref{fig:S2} presents housing-specific LLM correlations.

For each variable pair, both correlations were calculated using
the same matched observations. All ten variable pairs contributed
to the MAD for every LLM.
Let \(j\) and \(k\) index the five shared variables. For model \(m\),
\[
\mathrm{MAD}_{m}
=\frac{1}{10}
\sum_{1\leq j<k\leq 5}
\left|\rho_{jk,m}^{\mathrm{LLM}}-
\rho_{jk,m}^{\mathrm{CNEP+}}\right|.
\]

We used 1,000 stratified bootstrap resamples, retaining both
housing-condition outputs and the matched CNEP+ follow-up record
for each selected agent. The 95\% confidence intervals were
constructed using the 2.5th and 97.5th percentiles of the bootstrap
MAD estimates.

\begin{figure}[htbp]
\centering
\includegraphics[width=\linewidth]{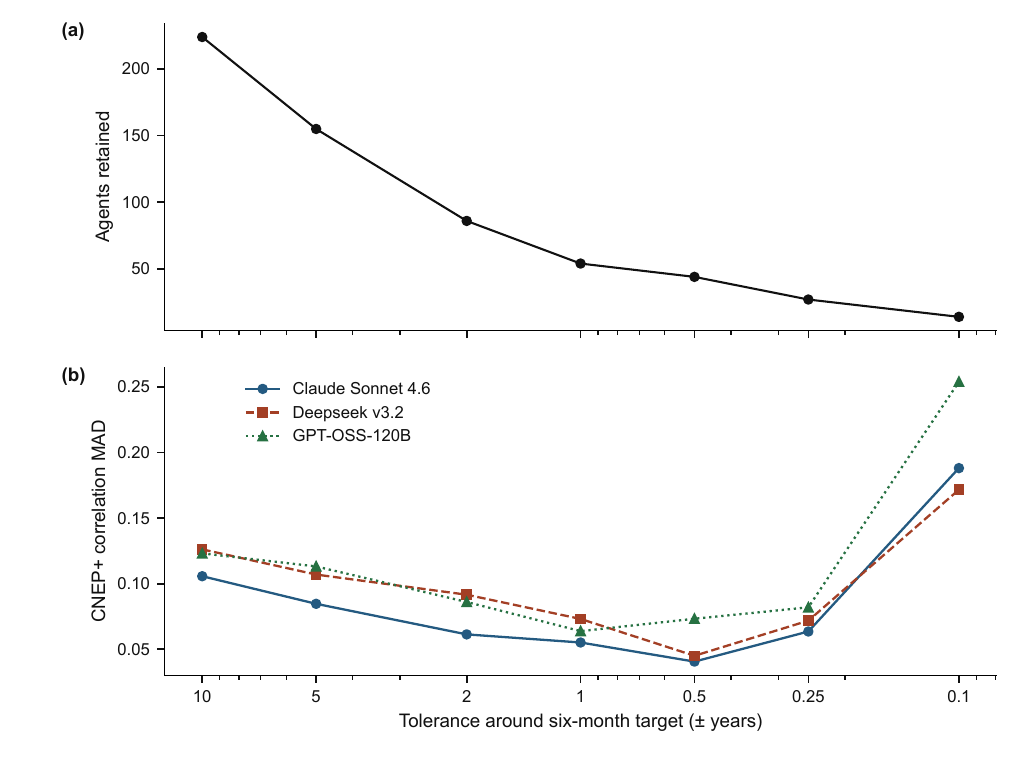}
\caption{Sensitivity of LLM--CNEP+ correlation MAD to follow-up
tolerance. Top, agents retained at each tolerance window; bottom,
model-specific MADs. Narrower windows change the included agents as
well as the permitted deviation from the six-month target.}
\label{fig:S3}
\end{figure}

Initial profiles were obtained from 224 CNEP+ agents who were HCV-susceptible at their earliest recorded activation event, which defined baseline. The baseline time for agent \(i\) is denoted by
\(t_i^0\). Eligible follow-up records had
susceptible, exposed, acute, chronic, or recovered HCV states. Among these eligible
records after activation, the record closest to the six-month target
was selected:
\[
r_i^*
=\operatorname*{arg\,min}_{r\in\mathcal{R}_i:\,t_{ir}>t_i^0}
\left|t_{ir}-(t_i^0+0.5)\right|,
\]
where \(\mathcal{R}_i\) denotes the eligible records and times in
years. No maximum deviation from the target was imposed during this selection.
To assess sensitivity to follow-up-record selection, we restricted
follow-up times to within $\pm$10, 5, 2, 1, 0.5, 0.25, and 0.1 years
of the six-month target. MAD varied across tolerance windows, although this variation may
also reflect changes in sample composition and the reduction in sample size from 224 to 14 agents (Figure~\ref{fig:S3}).

\FloatBarrier
\subsection{NYC HANES survey data}

The NYC HANES comparison used eight continuous variables: age, body
mass index, waist circumference, systolic blood pressure, diastolic
blood pressure, glycated hemoglobin (HbA1c), total cholesterol, and
high-density lipoprotein cholesterol. The four binary variables were
coded as female versus male, current daily or some-day smoking versus
no current smoking, ever street-drug use versus no use, and ever
injection-drug use versus no use. Unmapped or unknown binary responses
were treated as missing. Observed HCV antibody status was coded as
1 for positive and 0 for negative. LLMs were instructed to generate integer risk scores between 0 and 100.

For model \(m\), the NYC HANES MAD was
\[
\mathrm{MAD}_m
=\frac{1}{12}
\sum_{v=1}^{12}
\left|\rho_S(X_v,A)-\rho_S(X_v,L_m)\right|,
\]
where \(X_v\) denotes covariate \(v\), \(A\) is observed HCV antibody
status, \(L_m\) is the LLM-generated HCV risk score, and \(\rho_S\)
is Spearman's correlation. For each covariate, both correlations
were calculated using the same participants.

We used 1,000 bootstrap resamples stratified by HCV antibody status,
preserving 35 positive and 35 negative participants per resample.
The 95\% confidence intervals for both MAD and AUC were calculated using
the 2.5th and 97.5th percentiles of the bootstrap estimates.
Survey weights were not applied.

\FloatBarrier
\section{Knowledge-probe sensitivity to prompt wording}

Any change among increase, decrease, mixed, and uncertain direction
categories between prompt versions counted as a discrepancy. 
Claude exhibited no reversals between increase and decrease, while
DeepSeek and GPT-OSS each exhibited one reversal (Table~\ref{tab:S3}).

\begingroup
\small
\begin{longtable}[]{@{}
  >{\raggedright\arraybackslash}p{(\columnwidth - 6\tabcolsep) * \real{0.1813}}
  >{\raggedright\arraybackslash}p{(\columnwidth - 6\tabcolsep) * \real{0.3817}}
  >{\raggedright\arraybackslash}p{(\columnwidth - 6\tabcolsep) * \real{0.1908}}
  >{\raggedright\arraybackslash}p{(\columnwidth - 6\tabcolsep) * \real{0.2462}}@{}}
\caption{Variable pairs with different knowledge-probe direction classifications between the original and paraphrased prompts}\label{tab:S3}\\
\toprule\noalign{}
\begin{minipage}[b]{\linewidth}\raggedright
\textbf{Model}
\end{minipage} & \begin{minipage}[b]{\linewidth}\raggedright
\textbf{Variable pair}
\end{minipage} & \begin{minipage}[b]{\linewidth}\raggedright
\textbf{Original}
\end{minipage} & \begin{minipage}[b]{\linewidth}\raggedright
\textbf{Paraphrased}
\end{minipage} \\
\midrule\noalign{}
\endfirsthead
\toprule\noalign{}
\begin{minipage}[b]{\linewidth}\raggedright
\textbf{Model}
\end{minipage} & \begin{minipage}[b]{\linewidth}\raggedright
\textbf{Variable pair}
\end{minipage} & \begin{minipage}[b]{\linewidth}\raggedright
\textbf{Original}
\end{minipage} & \begin{minipage}[b]{\linewidth}\raggedright
\textbf{Paraphrased}
\end{minipage} \\
\midrule\noalign{}
\endhead
\bottomrule\noalign{}
\endlastfoot
Claude Sonnet 4.6 & Homelessness / Non-injection routes & decrease &
uncertain \\
DeepSeek v3.2 & Non-injection routes / HCV status & decrease &
increase \\
GPT-OSS-120B & Treatment / Non-injection routes & mixed & increase \\
GPT-OSS-120B & Non-injection routes / Overdose & increase & decrease \\
GPT-OSS-120B & HCV status / Overdose & uncertain & mixed \\
GPT-OSS-120B & Receptive sharing / Overdose & mixed & increase \\
\end{longtable}
\endgroup

\FloatBarrier
\section{Model execution details}

Table~\ref{tab:S4} summarizes the API identifiers and execution settings
specified in the generation notebooks. Requests used the Amazon Bedrock
Runtime Converse API through Boto3 in the \texttt{us-east-1} region,
with a single user message containing the relevant prompt. The analyses
used saved responses rather than issuing new model requests.

\begingroup
\small
\begin{table}[htbp]
\centering
\caption{Model configuration documented in the generation notebooks}
\label{tab:S4}
\begin{tabular}{@{}
  >{\raggedright\arraybackslash}p{(\textwidth - 4\tabcolsep) * \real{0.27}}
  >{\raggedright\arraybackslash}p{(\textwidth - 4\tabcolsep) * \real{0.61}}
  >{\centering\arraybackslash}p{(\textwidth - 4\tabcolsep) * \real{0.12}}@{}}
\toprule
\textbf{Model} & \textbf{Configured API identifier} & \textbf{Temperature} \\
\midrule
Claude Sonnet 4.6 & \path{us.anthropic.claude-sonnet-4-6} & 0 \\
DeepSeek v3.2 & \path{deepseek.v3.2} & 0 \\
GPT-OSS-120B & \path{openai.gpt-oss-120b-1:0} & 0 \\
\bottomrule
\multicolumn{3}{@{}p{\textwidth}@{}}{%
\footnotesize\textit{Note.} Identifiers are the configuration values in
the behavioral and knowledge-probe generation code. The survey
generation code specifies the same API, region, and temperature.
} \\
\end{tabular}
\end{table}
\endgroup
Outputs were requested in JSON format. Analysis validation checked for
duplicate identifiers and invalid outcome categories or numeric ranges.
Missing values were not imputed; comparison-specific handling is
described in the preceding sections.

{\small

}


\begin{thebibliography}{99}
\bibitem{main-ref31}
Niedzielewski, K., Bartczuk, R. P., Bielczyk, N., Bogucki, D., Dreger, F., Dudziuk, G., G{\'o}rski, {\L}., Gruziel-S{\l}omka, M., Haman, J., Kaczorek, A., Kisielewski, J., Krupa, B., Moszy{\'n}ski, A., Nowosielski, J. M., Radwan, M., Semeniuk, M., Tymoszuk, U., Zieli{\'n}ski, J., \& Rakowski, F. (2024). Forecasting SARS-CoV-2 epidemic dynamic in Poland with the pDyn agent-based model. \emph{Epidemics}, \emph{49}, Article 100801. \url{https://doi.org/10.1016/j.epidem.2024.100801}

\bibitem{main-ref17}
Gutfraind, A., Boodram, B., Prachand, N., Hailegiorgis, A., Dahari, H., \& Major, M. E. (2015). Agent-based model forecasts aging of the population of people who inject drugs in metropolitan Chicago and changing prevalence of hepatitis C infections. \emph{PLOS ONE}, \emph{10}(9), Article e0137993. \url{https://doi.org/10.1371/journal.pone.0137993}

\bibitem{main-ref13}
Eames, K., Bansal, S., Frost, S., \& Riley, S. (2015). Six challenges in measuring contact networks for use in modelling. \emph{Epidemics}, \emph{10}, 72--77. \url{https://doi.org/10.1016/j.epidem.2014.08.006}

\bibitem{main-ref24}
Lu, Y., Aleta, A., Du, C., Shi, L., \& Moreno, Y. (2024). LLMs and generative agent-based models for complex systems research. \emph{Physics of Life Reviews}, \emph{51}, 283--293. \url{https://doi.org/10.1016/j.plrev.2024.10.013}

\bibitem{main-ref2}
Adornetto, C., Mora, A., Hu, K., Izquierdo Garcia, L., Atchade-Adelomou, P., Greco, G., Alonso Pastor, L. A., \& Larson, K. (2025). Generative agents in agent-based modeling: Overview, validation, and emerging challenges. \emph{IEEE Transactions on Artificial Intelligence}, \emph{6}(12), 3165--3183. \url{https://doi.org/10.1109/TAI.2025.3566362}

\bibitem{main-ref34}
Williams, R., Hosseinichimeh, N., Majumdar, A., \& Ghaffarzadegan, N. (2023). \emph{Epidemic modeling with generative agents} [Preprint]. arXiv. \url{https://doi.org/10.48550/arXiv.2307.04986}

\bibitem{main-ref4}
Arum, C., Fraser, H., Artenie, A. A., Bivegete, S., Trickey, A., Alary, M., Astemborski, J., Iversen, J., Lim, A. G., MacGregor, L., Morris, M., Ong, J. J., Platt, L., Sack-Davis, R., van Santen, D. K., Solomon, S. S., Sypsa, V., Valencia, J., Van Den Boom, W., \ldots{} Vickerman, P. (2021). Homelessness, unstable housing, and risk of HIV and hepatitis C virus acquisition among people who inject drugs: A systematic review and meta-analysis. \emph{The Lancet Public Health}, \emph{6}(5), e309--e323. \url{https://doi.org/10.1016/S2468-2667(21)00013-X}



\bibitem{main-khezri2025}
Khezri, M., Kimball, S., McKnight, C., Rouhani, S., Bunting, A. M., Karamouzian, M., Ompad, D. C., \& Des Jarlais, D. (2025). Harms associated with injecting in public spaces: A global systematic review and meta-analysis. \emph{International Journal of Drug Policy}, \emph{140}, Article 104819. \url{https://doi.org/10.1016/j.drugpo.2025.104819}

\bibitem{main-ref6}
Ballard, A. M., Kesich, Z., Crane, H. M., Feinberg, J., Friedmann, P. D., Go, V. F., Jenkins, W. D., Korthuis, P. T., Miller, W. C., Pho, M. T., Seal, D. W., Smith, G. S., Stopka, T. J., Westergaard, R. P., Zule, W. A., Young, A. M., \& Cooper, H. L. F. (2025). Rural houselessness among people who use drugs in the United States: Results from the National Rural Opioid Initiative. \emph{Drug and Alcohol Dependence}, \emph{266}, Article 112498. \url{https://doi.org/10.1016/j.drugalcdep.2024.112498}

\bibitem{main-ref20}
Hotton, A., Mackesy-Amiti, M.-E., \& Boodram, B. (2021). Trends in homelessness and injection practices among young urban and suburban people who inject drugs: 1997--2017. \emph{Drug and Alcohol Dependence}, \emph{225}, Article 108797. \url{https://doi.org/10.1016/j.drugalcdep.2021.108797}

\bibitem{main-ref21}
Kan, M., Holm{\'e}n, E., Brynte, C., Ekstr{\"o}m, A. M., \& K{\aa}berg, M. (2026). Needle and syringe coverage per injection and factors associated with needle sharing among clients of the Stockholm Needle and Syringe Program, 2013--2023. \emph{International Journal of Drug Policy}, \emph{153}, Article 105299. \url{https://doi.org/10.1016/j.drugpo.2026.105299}

\bibitem{main-ref27}
Morris, M. D., Evans, J., Montgomery, M., Yu, M., Briceno, A., Page, K., \& Hahn, J. A. (2014). Intimate injection partnerships are at elevated risk of high-risk injecting: A multi-level longitudinal study of HCV-serodiscordant injection partnerships in San Francisco, CA. \emph{PLOS ONE}, \emph{9}(10), Article e109282. \url{https://doi.org/10.1371/journal.pone.0109282}

\bibitem{main-ref28}
Morris, M. D., Yen, I. H., Shiboski, S., Evans, J. L., \& Page, K. (2020). Housing stability and hepatitis C infection for young adults who inject drugs: Examining the relationship of consistent and intermittent housing status on HCV infection risk. \emph{Journal of Urban Health}, \emph{97}(6), 831--844. \url{https://doi.org/10.1007/s11524-020-00445-7}

\bibitem{main-ref14}
Gahrton, C., Nav{\'e}r, G., Warnqvist, A., Dalgard, O., Aleman, S., \& K{\aa}berg, M. (2024). Changes in hepatitis C virus prevalence and incidence among people who inject drugs in the direct acting antiviral era. \emph{International Journal of Drug Policy}, \emph{128}, Article 104433. \url{https://doi.org/10.1016/j.drugpo.2024.104433}

\bibitem{main-ref7}
Byrne, T., Troszak, L., Midboe, A. M., Fincke, B. G., Shwartz, M., Gifford, A. L., \& McInnes, D. K. (2019). A novel measure to assess variation in hepatitis C prevalence among homeless and unstably housed veterans, 2011--2016. \emph{Public Health Reports}, \emph{134}(2), 126--131. \url{https://doi.org/10.1177/0033354918821071}

\bibitem{main-ref18}
Hahn, J. A., Page-Shafer, K., Lum, P. J., Bourgois, P., Stein, E., Evans, J. L., Busch, M. P., Tobler, L. H., Phelps, B., \& Moss, A. R. (2002). Hepatitis C virus seroconversion among young injection drug users: Relationships and risks. \emph{The Journal of Infectious Diseases}, \emph{186}(11), 1558--1564. \url{https://doi.org/10.1086/345554}

\bibitem{main-ref12}
Dunne, N., Casey, M.-B., \& Ivers, J.-H. (2025). What are the risk factors of non-fatal overdose among persons who use opioids? A systematic review and meta-analysis. \emph{Journal of Substance Use and Addiction Treatment}, \emph{171}, Article 209630. \url{https://doi.org/10.1016/j.josat.2025.209630}

\bibitem{main-ref26}
Marcus, R., Cha, S., Sionean, C., Kanny, D., \& National HIV Behavioral Surveillance Study Group. (2022). HIV injection risk behaviors among HIV-negative people who inject drugs experiencing homelessness, 23 U.S. cities. \emph{Journal of Social Distress and Homelessness}, \emph{31}(1), 96--104. \url{https://doi.org/10.1080/10530789.2021.1892931}

\bibitem{main-ref16}
Gunaratne, M. P., Loeb, T. A., McFall, A. M., Ganapathi, L., Baishya, J. J., Kedar, A., Sinha, A. K., Srikrishnan, A. K., Solomon, S. S., Lucas, G. M., \& Mehta, S. H. (2026). Characterizing non-fatal overdose among people who inject drugs in India: Prevalence and risk factor analysis, 2022--2024. \emph{International Journal of Drug Policy}, \emph{147}, Article 105073. \url{https://doi.org/10.1016/j.drugpo.2025.105073}

\bibitem{main-ref15}
Guarino, H., Frank, D., Quinn, K., Kim, D., Gile, K., Ruggles, K., Friedman, S. R., \& Mateu-Gelabert, P. (2023). Syndemic factors associated with non-fatal overdose among young opioid users in New York City. \emph{Frontiers in Public Health}, \emph{11}, Article 1195657. \url{https://doi.org/10.3389/fpubh.2023.1195657}

\bibitem{main-ref5}
Austin, A. E., Shiue, K. Y., Naumann, R. B., Figgatt, M. C., Gest, C., \& Shanahan, M. E. (2021). Associations of housing stress with later substance use outcomes: A systematic review. \emph{Addictive Behaviors}, \emph{123}, Article 107076. \url{https://doi.org/10.1016/j.addbeh.2021.107076}

\bibitem{main-ref11}
Degenhardt, L., Webb, P., Colledge-Frisby, S., Ireland, J., Wheeler, A., Ottaviano, S., Willing, A., Kairouz, A., Cunningham, E. B., Hajarizadeh, B., Leung, J., Tran, L. T., Price, O., Peacock, A., Vickerman, P., Farrell, M., Dore, G. J., Hickman, M., \& Grebely, J. (2023). Epidemiology of injecting drug use, prevalence of injecting-related harm, and exposure to behavioural and environmental risks among people who inject drugs: A systematic review. \emph{The Lancet Global Health}, \emph{11}(5), e659--e672. \url{https://doi.org/10.1016/S2214-109X(23)00057-8}

\bibitem{main-ref22}
Karandinos, G., Unick, J., Ondocsin, J., Holm, N., Mars, S., Montero, F., Rosenblum, D., \& Ciccarone, D. (2024). Decrease in injection and rise in smoking and snorting of heroin and synthetic opioids, 2000--2021. \emph{Drug and Alcohol Dependence}, \emph{263}, Article 111419. \url{https://doi.org/10.1016/j.drugalcdep.2024.111419}

\bibitem{main-ref32}
Rahimi-Movaghar, A., Amin-Esmaeili, M., Shadloo, B., Noroozi, A., \& Malekinejad, M. (2015). Transition to injecting drug use in Iran: A systematic review of qualitative and quantitative evidence. \emph{International Journal of Drug Policy}, \emph{26}(9), 808--819. \url{https://doi.org/10.1016/j.drugpo.2015.04.018}

\bibitem{main-ref29}
New York City Department of Health and Mental Hygiene, \& United States Department of Health and Human Services, Centers for Disease Control and Prevention, National Center for Health Statistics. (2011). \emph{New York City Health and Nutrition Examination Survey (NYC HANES), 2004} (ICPSR 31421; Version 1) [Data set]. Inter-university Consortium for Political and Social Research. \url{https://doi.org/10.3886/ICPSR31421.v1}



\bibitem{main-ref19}
Nasution, B. I., Nugraha, Y., Aminanto, M. E., \& Kanggrawan, J. I. (2026). Synthetic data publishing using Bayesian network: A case study of COVID-19 patients data in Jakarta, Indonesia. \emph{Journal of Computational Social Science}, \emph{9}, Article 70. \url{https://doi.org/10.1007/s42001-026-00501-3}

\bibitem{main-ref3}
Anderson, T., Von Hoene, S., Cinar, O. Y., Von Hoene, E., Roess, A., Crooks, A., \& Kavak, H. (2026). \emph{Using zero-shot LLM-generated survey data for geographically explicit population synthesis} [Preprint]. arXiv. \url{https://doi.org/10.48550/arXiv.2605.27401}

\bibitem{main-ref8}
Zhao, W. X., Zhou, K., Li, J., Tang, T., Dong, Z., Hou, Y., Zhang, B., Min, Y., Zhang, J., Liu, P., Wang, X., Du, Y., Yang, C., Chen, Y., Chen, Z., Jiang, J., Ren, R., Li, Y., Tang, X., \ldots{} Wen, J.-R. (2026). A survey of large language models. \emph{Frontiers of Computer Science}, \emph{20}(12), Article 2012627. \url{https://doi.org/10.1007/s11704-026-60308-3}

\bibitem{main-ref1}
Abuelezam, N. N., Rough, K., \& Seage, G. R., III. (2013). Individual-based simulation models of HIV transmission: Reporting quality and recommendations. \emph{PLOS ONE}, \emph{8}(9), Article e75624. \url{https://doi.org/10.1371/journal.pone.0075624}

\bibitem{main-ref30}
Rafikova, A., \& Voronin, A. (2026). ChatGPT as a research proxy: Simulating human attitudes in social science research. \emph{Journal of Computational Social Science}, \emph{9}, Article 17. \url{https://doi.org/10.1007/s42001-025-00452-1}

\end{thebibliography}

\begin{thebibliography}{99}
\interlinepenalty=10000\relax
\bibitem{supplement-ref15}
Khezri, M., Kimball, S., McKnight, C., Rouhani, S., Bunting, A. M., Karamouzian, M., Ompad, D. C., \& Des Jarlais, D. (2025). Harms associated with injecting in public spaces: A global systematic review and meta-analysis. \emph{International Journal of Drug Policy}, \emph{140}, Article 104819. \url{https://doi.org/10.1016/j.drugpo.2025.104819}

\bibitem{supplement-ref3}
Eckhardt, B., Winkelstein, E. R., Shu, M. A., Carden, M. R., McKnight, C., Des Jarlais, D. C., Glesby, M. J., Marks, K., \& Edlin, B. R. (2017). Risk factors for hepatitis C seropositivity among young people who inject drugs in New York City: Implications for prevention. \emph{PLOS ONE}, \emph{12}(5), Article e0177341. \url{https://doi.org/10.1371/journal.pone.0177341}

\bibitem{supplement-ref9}
Hahn, J. A., Page-Shafer, K., Lum, P. J., Bourgois, P., Stein, E., Evans, J. L., Busch, M. P., Tobler, L. H., Phelps, B., \& Moss, A. R. (2002). Hepatitis C virus seroconversion among young injection drug users: Relationships and risks. \emph{The Journal of Infectious Diseases}, \emph{186}(11), 1558--1564. \url{https://doi.org/10.1086/345554}

\bibitem{supplement-ref20}
Pouget, E. R., Hagan, H., \& Des Jarlais, D. C. (2012). Meta-analysis of hepatitis C seroconversion in relation to shared syringes and drug preparation equipment. \emph{Addiction}, \emph{107}(6), 1057--1065. \url{https://doi.org/10.1111/j.1360-0443.2011.03765.x}

\bibitem{supplement-ref4}
Gahrton, C., Navér, G., Warnqvist, A., Dalgard, O., Aleman, S., \& Kåberg, M. (2024). Changes in hepatitis C virus prevalence and incidence among people who inject drugs in the direct acting antiviral era. \emph{International Journal of Drug Policy}, \emph{128}, Article 104433. \url{https://doi.org/10.1016/j.drugpo.2024.104433}

\bibitem{supplement-ref18}
Mackesy-Amiti, M. E., Boodram, B., Page, K., \& Latkin, C. (2023). Injection partnership characteristics and HCV status associations with syringe and equipment sharing among people who inject drugs. \emph{BMC Public Health}, \emph{23}(1), Article 1191. \url{https://doi.org/10.1186/s12889-023-16133-5}

\bibitem{supplement-ref12}
Kan, M., Holmén, E., Brynte, C., Ekström, A. M., \& Kåberg, M. (2026). Needle and syringe coverage per injection and factors associated with needle sharing among clients of the Stockholm Needle and Syringe Program, 2013--2023. \emph{International Journal of Drug Policy}, \emph{153}, Article 105299. \url{https://doi.org/10.1016/j.drugpo.2026.105299}

\bibitem{supplement-ref11}
Jenkins, L. M., Banta-Green, C. J., Maynard, C., Kingston, S., Hanrahan, M., Merrill, J. O., \& Coffin, P. O. (2011). Risk factors for nonfatal overdose at Seattle-area syringe exchanges. \emph{Journal of Urban Health}, \emph{88}(1), 118--128. \url{https://doi.org/10.1007/s11524-010-9525-6}

\bibitem{supplement-ref6}
Guarino, H., Frank, D., Quinn, K., Kim, D., Gile, K., Ruggles, K., Friedman, S. R., \& Mateu-Gelabert, P. (2023). Syndemic factors associated with non-fatal overdose among young opioid users in New York City. \emph{Frontiers in Public Health}, \emph{11}, Article 1195657. \url{https://doi.org/10.3389/fpubh.2023.1195657}

\bibitem{supplement-ref2}
Dunne, N., Casey, M.-B., \& Ivers, J.-H. (2025). What are the risk factors of non-fatal overdose among persons who use opioids? A systematic review and meta-analysis. \emph{Journal of Substance Use and Addiction Treatment}, \emph{171}, Article 209630. \url{https://doi.org/10.1016/j.josat.2025.209630}

\bibitem{supplement-walker2025}
Walker, J. G., Akiyama, M. J., Artenie, A., Cleland, C. M., Lizcano, J. A., Musyoki, H., Nyakowa, M., Cherutich, P., Kurth, A. E., \& Vickerman, P. (2025). Impact of scaling up harm reduction interventions on injecting risk behaviours, ART outcomes and HIV incidence among people who inject drugs in Kenya. \emph{International Journal of Drug Policy}, \emph{140}, Article 104824. \url{https://doi.org/10.1016/j.drugpo.2025.104824}

\bibitem{supplement-ref14}
Karlsson, N., Kåberg, M., Berglund, T., Hammarberg, A., Widman, L., \& Ekström, A. M. (2021). A prospective cohort study of risk behaviours, retention and loss to follow-up over 5 years among women and men in a needle exchange program in Stockholm, Sweden. \emph{International Journal of Drug Policy}, \emph{90}, Article 103059. \url{https://doi.org/10.1016/j.drugpo.2020.103059}

\bibitem{supplement-dong2020}
Dong, H., Hayashi, K., Milloy, M.-J., DeBeck, K., Singer, J., Wong, H., Wood, E., \& Kerr, T. (2020). Changes in substance use in relation to opioid agonist therapy among people who use drugs in a Canadian setting. \emph{Drug and Alcohol Dependence}, \emph{212}, Article 108005. \url{https://doi.org/10.1016/j.drugalcdep.2020.108005}

\bibitem{supplement-ref7}
Gunaratne, M. P., Loeb, T. A., McFall, A. M., Ganapathi, L., Baishya, J. J., Kedar, A., Sinha, A. K., Srikrishnan, A. K., Solomon, S. S., Lucas, G. M., \& Mehta, S. H. (2026). Characterizing non-fatal overdose among people who inject drugs in India: Prevalence and risk factor analysis, 2022--2024. \emph{International Journal of Drug Policy}, \emph{147}, Article 105073. \url{https://doi.org/10.1016/j.drugpo.2025.105073}

\bibitem{supplement-ref10}
Des Jarlais, D., Weng, C. A., Feelemyer, J., \& McKnight, C. (2023). Non-fatal drug overdose among persons who inject drugs during first two years of the COVID-19 pandemic in New York City: Prevalence, risk factors. \emph{Drug and Alcohol Dependence Reports}, \emph{8}, Article 100171. \url{https://doi.org/10.1016/j.dadr.2023.100171}

\bibitem{supplement-ref21}
Rafful, C., Orozco, R., Rangel, G., Davidson, P., Werb, D., Beletsky, L., \& Strathdee, S. A. (2018). Increased non-fatal overdose risk associated with involuntary drug treatment in a longitudinal study with people who inject drugs. \emph{Addiction}, \emph{113}(6), 1056--1063. \url{https://doi.org/10.1111/add.14159}

\bibitem{supplement-ref8}
Hahn, J. A., Evans, J. L., Davidson, P. J., Lum, P. J., \& Page, K. (2010). Hepatitis C virus risk behaviors within the partnerships of young injecting drug users. \emph{Addiction}, \emph{105}(7), 1254--1264. \url{https://doi.org/10.1111/j.1360-0443.2010.02949.x}

\bibitem{supplement-ref27}
Tran, L. T., Peacock, A., Colledge, S., Memedovic, S., Grebely, J., Leung, J., Larney, S., Trickey, A., Stone, J., Vickerman, P., Hickman, M., \& Degenhardt, L. (2020). Injecting risk behaviours amongst people who inject drugs: A global multi-stage systematic review and meta-analysis. \emph{International Journal of Drug Policy}, \emph{84}, Article 102866. \url{https://doi.org/10.1016/j.drugpo.2020.102866}

\bibitem{supplement-ref22}
Rahimi-Movaghar, A., Amin-Esmaeili, M., Shadloo, B., Noroozi, A., \& Malekinejad, M. (2015). Transition to injecting drug use in Iran: A systematic review of qualitative and quantitative evidence. \emph{International Journal of Drug Policy}, \emph{26}(9), 808--819. \url{https://doi.org/10.1016/j.drugpo.2015.04.018}

\bibitem{supplement-ref16}
Kral, A. H., Lambdin, B. H., Browne, E. N., Wenger, L. D., Bluthenthal, R. N., Zibbell, J. E., \& Davidson, P. J. (2021). Transition from injecting opioids to smoking fentanyl in San Francisco, California. \emph{Drug and Alcohol Dependence}, \emph{227}, Article 109003. \url{https://doi.org/10.1016/j.drugalcdep.2021.109003}

\bibitem{supplement-desjarlais2018}
Des Jarlais, D. C., Arasteh, K., Feelemyer, J., McKnight, C., Barnes, D. M., Perlman, D. C., Uuskula, A., Cooper, H. L. F., \& Tross, S. (2018). Hepatitis C virus prevalence and estimated incidence among new injectors during the opioid epidemic in New York City, 2000--2017: Protective effects of non-injecting drug use. \emph{Drug and Alcohol Dependence}, \emph{192}, 74--79. \url{https://doi.org/10.1016/j.drugalcdep.2018.07.034}

\bibitem{supplement-ref19}
Morris, M. D., Evans, J., Montgomery, M., Yu, M., Briceno, A., Page, K., \& Hahn, J. A. (2014). Intimate injection partnerships are at elevated risk of high-risk injecting: A multi-level longitudinal study of HCV-serodiscordant injection partnerships in San Francisco, CA. \emph{PLOS ONE}, \emph{9}(10), Article e109282. \url{https://doi.org/10.1371/journal.pone.0109282}

\bibitem{supplement-ref23}
Rudolph, A. E., \& Rhodes, S. E. (2023). Network correlates of using a syringe after an injection partner among women who inject drugs in Philadelphia Pennsylvania. \emph{AIDS and Behavior}, \emph{27}(3), 957--968. \url{https://doi.org/10.1007/s10461-022-03832-0}

\bibitem{supplement-ref25}
Spelman, T., Sacks-Davis, R., Dietze, P., Higgs, P., \& Hellard, M. (2019). Injection drug network characteristics as a predictor of injection behaviour. \emph{Epidemiology and Infection}, \emph{147}, Article e173. \url{https://doi.org/10.1017/S095026881900061X}

\bibitem{supplement-ref24}
Schiavon, S., Hodgin, K., Sellers, A., Word, M., Galbraith, J. W., Dantzler, J., \& Cropsey, K. L. (2018). Medical, psychosocial, and treatment predictors of opioid overdose among high risk opioid users. \emph{Addictive Behaviors}, \emph{86}, 51--55. \url{https://doi.org/10.1016/j.addbeh.2018.05.029}

\end{thebibliography}
\end{document}